\documentclass{article}
\usepackage[final]{nejlt}

\usepackage{booktabs}
\usepackage{graphicx}
\usepackage{array}
\usepackage{url}
\usepackage{xurl} % break long URLs at any character; prevents the Sec. 3.3
\title{Evaluation design conditions the expert-vs-auto MeSH gap: a controlled comparison of bag-of-words and BiomedBERT on the Cohen benchmark}



\author{
Samuel M. Okoe-Mensah, Accra, Ghana {\tt \small sammy.okmens@gmail.com}
}

\begin{document}

\abstract{A systematic review begins with someone reading thousands of abstracts to identify the few that are relevant, and classifiers are used to prioritise that reading. Their inputs are often augmented with Medical Subject Headings (MeSH), assigned either by expert indexers weeks or months after publication or by automatic tools at once. We did not identify prior work comparing the two directly as classifier features, or asking whether that comparison's outcome depends on how the classifier is evaluated. We compare expert assignment against one mechanical procedure, substring matching against a MeSH vocabulary drawn from the benchmark; the results bound that procedure rather than automatic MeSH indexing in general. Using the \newcite{cohen2006drug} drug-class benchmark on three topics, we characterise a bag-of-words logistic regression classifier (seven reruns) and BiomedBERT (five seeds), then examine how the Statins result changes under alternative evaluation designs. Under the canonical 5-fold full-corpus design, the bag-of-words expert-vs-auto gap on Statins is $+0.096$ WSS@95\%. Measured against each mode's own no-MeSH baseline, the expert augmentation contributes $+0.084$ and the mechanical augmentation $-0.006$ with an interval spanning zero. Stratified subsampling to matched corpus size ($n=803$) reduces the gap by roughly two thirds, to $+0.033$ (95\% bootstrap CI includes zero); 10-fold cross-validation at full corpus size reduces it by roughly four fifths, to $+0.021$ (CI excludes zero by a small margin). BiomedBERT under canonical evaluation gives $+0.020$, a difference of $0.001$ from the bag-of-words 10-fold result. An empirical power analysis on a single canonical 5-fold run per topic indicates that a Statins-sized effect at Opioids' or ADHD's per-fold variance would not have been detectable at that design ($\text{MDE}_{\text{Opioids}} = 0.254$, $\text{MDE}_{\text{ADHD}} = 0.384$); at the pooled fold count of the multi-run protocol the bound depends on an effective sample size the design does not determine. Of the designs tested, the canonical one is where the gap is most cleanly detected, and the gap attenuates under designs that increase per-fold training volume. A representation asymmetry remains: 15.1\% of Statins inputs exceed BiomedBERT's 512-token cap in the expert-MeSH mode, and truncation preferentially removes the appended MeSH terms. This could attenuate the transformer's gap but cannot be isolated from the training-volume effect without a no-truncation comparison. For screening pipelines using transformer classifiers or 10-fold bag-of-words, the observed expert-vs-auto gap on the topics tested is approximately $0.02$ WSS@95\%, with confidence intervals that either include zero or exclude it only narrowly. More broadly, benchmark conclusions about feature sources can change substantially under reasonable changes to the evaluation design.}

\maketitle

\section{Introduction}

Systematic reviews assemble published evidence relevant to a clinical question. Screening reviewers read thousands of titles and abstracts to identify the small subset relevant to the review; the work is expensive and slow, and automated screening tools have been studied since at least \newcite{cohen2006drug}. Their drug-class benchmark established a 15-topic testbed and an evaluation metric, work saved over sampling at 95\% recall (WSS@95\%), that the field continues to use. The standard input representation for screening classifiers has been an article's title and abstract, often augmented with MeSH headings drawn from the National Library of Medicine's controlled vocabulary. The augmentation captures topic information the abstract may not state explicitly, such as the indication that an article is about a specific drug class even when no specific drug name appears in the title.

MeSH terms are produced by two mechanisms. Professional indexers at the National Library of Medicine assign MeSH headings to each PubMed-indexed article based on its full content; these expert annotations are the canonical reference. Automated tools separately assign MeSH terms mechanically through string matching, concept linking, or model-based prediction. Whether the two annotation sources are interchangeable as classifier features has practical importance: expert annotation is delayed by weeks or months after publication, while mechanical assignment is instantaneous and can be applied to preprints or recently-published articles that lack expert indexing. If the two produce comparable downstream classifier performance, mechanical assignment is a viable substitute. If they do not, systematic review screening depends on annotation latency in a way that limits coverage of recent literature.

We did not identify a prior study that isolates annotation mechanism while holding MeSH vocabulary constant. Existing work treats expert MeSH as ground truth and auto-MeSH prediction as an algorithmic problem in its own right \cite{mao2017meshnow,jin2018pico}. Studies that augment classifier inputs with MeSH features typically use the expert-assigned version without varying the assignment mechanism \cite{cohen2008,matwin2010}. The nearest analogue concerns features more broadly: \newcite{scott1999} compared phrases, synonyms, and hypernyms as features for Reuters text classification. We likewise did not find the interaction with evaluation design raised; the canonical 5-fold cross-validation at native topic size is treated throughout this literature as a methodological default rather than an experimental variable.

This paper takes up both questions through a controlled experiment on three Cohen topics, holding the MeSH vocabulary constant and varying the assignment mechanism. Two classifier families serve as the test platform: a bag-of-words logistic regression classifier and BiomedBERT (formerly PubMedBERT) \cite{gu2021biomedbert}, a domain-pretrained contextual transformer. We characterise each topic-classifier combination as a distribution --- seven-rerun for bag-of-words, five-seed for BiomedBERT --- and supplement the main comparison with two robustness analyses on the Statins finding (subsampling to matched corpus size; 10-fold cross-validation at full corpus size) and an empirical power analysis on the per-topic variances. Bag-of-words is used here as a controlled testbed for isolating a mechanism, not as a competing system for current screening practice; Section~\ref{sec:related} sets out why the mechanism question is independent of which classifier performs best.

Our research questions and paired hypotheses are:

\begin{itemize}

\item \textbf{RQ1 / H1.} Does mechanical MeSH assignment by substring matching produce comparable WSS@95\% to expert MeSH assignment as classifier features, holding vocabulary and classifier constant? \textit{H1: expert MeSH produces a positive advantage under the canonical design on Statins.}
\item \textbf{RQ2 / H2.} Does the answer depend on the classifier? \textit{H2: BiomedBERT reduces the expert-vs-auto gap relative to bag-of-words at Statins.}
\item \textbf{RQ3 / H3.} Does the answer depend on the evaluation design? \textit{H3: the Statins expert-vs-auto gap magnitude is conditional on per-fold training volume and per-topic corpus size.}
\end{itemize}

Under the canonical design the expected Statins advantage appears; BiomedBERT attenuates it, and the bag-of-words advantage itself shrinks under both robustness analyses to magnitudes close to the BiomedBERT result. We do not adjudicate whether the transformer effect is architecturally distinct from the design-sensitivity effect; the data are consistent with a common training-volume mechanism affecting both. At Opioids and ADHD the gap is not detectable under any design we tested. Section~\ref{sec:design-sensitivity} sets out what a single canonical run could have detected at those topics' per-fold variances, and why the corresponding bound at the pooled fold count of the multi-run protocol cannot be settled here.

We make three contributions.

\begin{itemize}

\item \textbf{A design-conditional characterisation of a widely-cited screening finding.} We show that the Statins expert-vs-auto MeSH gap on the Cohen benchmark attenuates by roughly two thirds under matched-corpus-size subsampling and by roughly four fifths under 10-fold cross-validation, so that the canonical-design magnitude is sensitive both to corpus size and to fold configuration.
\item \textbf{A controlled cross-classifier comparison of annotation mechanism.} Holding vocabulary and evaluation constant, we compare bag-of-words and BiomedBERT on three Cohen drug-class topics, treating each topic-classifier combination as a distribution rather than a point estimate through seven-rerun and five-seed multi-run protocols with archived per-fold values.

\item \textbf{A power-analytic reading of the cross-topic null.} At the single canonical 5-fold run the power analysis characterises, the empirical minimum detectable effect at Opioids' and ADHD's per-fold variances exceeds the Statins effect size, so that null does not distinguish absence of a mechanism effect from absence of detection power. At the pooled fold count of the multi-run protocol the same calculation gives a smaller bound, and which applies depends on an effective sample size the design does not determine. We report this as a design limitation rather than a substantive finding.
\end{itemize}

\section{Related work}\label{sec:related}

\newcite{cohen2006drug} introduced the drug-class benchmark of 15 systematic review topics and reported WSS@95\% values for a voting perceptron classifier with binary bag-of-words features. Their cross-topic average was approximately 18.5\%, with four topics showing WSS@95\% below 5\%. Subsequent work has continued to use this benchmark. \newcite{cohen2008} reported SVM-based results with n-gram features and MeSH terms, using AUC rather than WSS@95\%. \newcite{matwin2010} used factorised complement naive Bayes and reported WSS@95\% improvements on several topics. Across this body of work, MeSH augmentation appears as a feature source but its assignment mechanism is treated as fixed, and the 5-fold cross-validation at native topic size is treated as a methodological default rather than an experimental variable.

Recent screening work has shifted toward transformer and large-language-model classifiers, with reported performance for GPT-class systems on systematic review tasks that often exceeds earlier statistical baselines \cite{guo2024llm,alshami2023chatgpt,wang2024boolean}. That shift does not displace the question asked here. Whether two annotation sources are interchangeable as classifier inputs is separable from which input pipeline produces the highest screening performance overall, and a simpler classifier isolates the first question more cleanly than a stronger one would.

The reliability of biomedical NLP benchmarks under label variation has received recent attention \cite{plank2022}; the present paper extends this concern to evaluation-design variation. We show that a widely-cited benchmark result is stable under some perturbations of the canonical evaluation and unstable under others, and that this instability has been invisible because the field has retained the canonical design as a default.

MeSH-based feature augmentation appears in the broader text classification literature as one instance of ontology-driven feature engineering. The semasiological versus onomasiological distinction in lexical semantics \cite{baldinger1980,geeraerts2010}, with origins in \newcite{saussure1959}, describes two directions of vocabulary access, and distributional semantics \cite{harris1954distributional,sahlgren2008,lenci2018} provides the theoretical underpinning for bag-of-words methods. Whether that distinction maps onto the mechanical-versus-expert contrast is one interpretive frame for the canonical-design Statins finding; because the present experiments do not test it, we defer it to Section~\ref{sec:lexical}.

Biomedical transformer models \cite{gu2021biomedbert,lee2020biobert,yasunaga2022linkbert} are pretrained on PubMed abstracts and have produced state-of-the-art results on biomedical NLP benchmarks including the BLURB suite. We use BiomedBERT (rather than BioBERT, BioLinkBERT, or ClinicalBERT) because BLURB benchmarking placed BiomedBERT-base above alternatives at comparable parameter count \cite{gu2021biomedbert} and because our comparison is not intended to identify the strongest screening architecture; we need a domain-pretrained contextual transformer with tractable fine-tuning cost. Their use for systematic review screening has been less studied than their use for biomedical question answering or named entity recognition. We use BiomedBERT as a bridge from the controlled bag-of-words experiment to representations that are closer to current screening practice.

\section{Methods}

\subsection{Benchmark and topic selection}

We use three drug-class topics from the \newcite{cohen2006drug} benchmark: Statins, Opioids, and ADHD. The benchmark provides PubMed identifiers and inclusion labels from completed systematic reviews. Article texts and MeSH terms were retrieved through the NCBI Entrez API and cached locally to ensure reproducibility. Table~\ref{tab:topics} summarises topic characteristics.

\begin{table}[htbp]
\centering
\small
\begin{tabular}{lrrr}
\toprule
Topic & Total articles & Included & Inclusion rate \\
\midrule
Statins & 2,744 & 152 & 5.5\% \\
Opioids & 1,772 & 43 & 2.4\% \\
ADHD & 803 & 83 & 10.3\% \\
\bottomrule
\end{tabular}
\caption{Cohen drug-class topics used in this study. Counts are the post-retrieval subset described in the text, not the benchmark totals.}
\label{tab:topics}
\end{table}

The benchmark records two triage decisions per article, at abstract level and at article level, and we use the abstract-level decision as the inclusion label. Its coding scheme distinguishes inclusion from several kinds of exclusion: \texttt{I} marks an included record, \texttt{E} a non-specifically excluded one, and the numerals 1 to 9 exclusions that carry a stated reason, such as wrong population or wrong publication type. All non-\texttt{I} codes are exclusions, and we treat them as such. The reason-coded exclusions are not evenly distributed across topics: 190 of the 1,915 Opioids records and 81 of the 851 ADHD records carry a numeral, against 1 of the 3,465 Statins records. Abstract-level inclusions number 173 for Statins, 48 for Opioids and 84 for ADHD; article-level inclusions number 85, 15 and 20 and are a subset of them, which is why our counts differ from the article-level figures more commonly quoted for this benchmark.

The benchmark lists 3,465 Statins, 1,915 Opioids and 851 ADHD records. Table~\ref{tab:topics} reports the subset for which an abstract was retrievable through Entrez at the time of collection; records without a retrievable abstract were dropped, which accounts for the difference. That attrition removed included articles as well as excluded ones, taking the abstract-level inclusion counts from 173, 48 and 84 to the 152, 43 and 83 that Table~\ref{tab:topics} reports, so the inclusion rates in the table are those of the retrieved subset rather than of the benchmark.

The three topics span different inclusion rates and different clinical domains (cardiovascular, analgesic, behavioural). The Statins topic is the most studied in the screening literature and has been used as the primary reference in several follow-up papers. Opioids and ADHD differ from Statins in both corpus size and inclusion rate, providing cross-topic variation on dimensions that prior work has shown to affect classifier performance \cite{cohen2008,matwin2010}. The Cohen benchmark structure assigns each topic its native size, and the literature has retained this convention; we follow it in the main comparison and explicitly relax it in the robustness analysis (Section~\ref{sec:robustness}).

\subsection{Text modes}

Each article is represented in four alternative input forms:

\begin{enumerate}
\item \texttt{abstract}: the abstract text alone.
\item \texttt{title\_abstract}: the article title concatenated with the abstract.
\item \texttt{title\_abstract\_mesh}: title plus abstract plus expert-assigned MeSH terms from the PubMed indexer, concatenated as space-separated tokens.
\item \texttt{auto\_mesh}: abstract plus mechanically-assigned MeSH terms, where the assignment uses case-insensitive substring matching of MeSH headings against a vocabulary built from the union of all MeSH terms of at least four characters observed across the cached benchmark records.
\end{enumerate}

Modes 1 and 2 establish baseline performance without MeSH augmentation. Modes 3 and 4 instantiate the central manipulation: both augment the text with MeSH vocabulary, but mode 3 uses the expert assignment and mode 4 uses the mechanical assignment.

Two properties of this construction need stating. First, the mode 4 vocabulary is drawn from the benchmark cache rather than from the full MeSH thesaurus, so both modes draw on the same body of MeSH terms and differ in how those terms are attached to an article. The cache spans more of the benchmark than the three topics studied here (4,740 terms from 6,216 cached records, against 5,319 articles in the three topics), so the vocabulary is larger than a topic-scoped one would be and smaller than the full thesaurus. It is, precisely, the set of MeSH headings an indexer assigned somewhere in the benchmark. A matcher deployed against the full thesaurus would work from a larger vocabulary and a fold-scoped one from a smaller, and the vocabulary used here therefore admits matches the indexer did not assign for a given article while excluding headings that appear nowhere in the benchmark. Section~\ref{sec:assignment} reports the rate of unassigned matches; we do not establish which way that construction moves WSS@95\%, and Section~\ref{sec:limitations} returns to the choice.

Second, mode 4 appends its terms to the abstract alone rather than to title plus abstract, following the construction used in the earlier work from which the pipeline derives. Modes 3 and 4 therefore differ in the presence of the title as well as in the assignment mechanism. Section~\ref{sec:decomposition} measures each augmentation against its own no-MeSH baseline, which removes the title from the comparison.

\subsection{Classifiers}\label{sec:classifiers}

The bag-of-words pipeline tokenises with a Keras Tokenizer using stop-word removal and unigram binary features, feeds these into a regularised logistic regression classifier, and chooses the L2 regularisation strength by cross-validation on the training folds. We report the regularised variant throughout. Full pipeline implementation appears in the project repository.

The BiomedBERT pipeline uses \url{microsoft/BiomedNLP-BiomedBERT-base-uncased-abstract}, the abstract-pretrained variant of BiomedBERT (formerly PubMedBERT) \cite{gu2021biomedbert}. Hyperparameters were chosen as BERT-fine-tuning defaults rather than topic-tuned: learning rate 2e-5, batch size 16, 3 fine-tuning epochs, maximum sequence length 512 tokens. Token-length analysis of the input corpus shows that 4.6--7.9\% of abstracts across the three topics exceed this 512-token cap in the \texttt{abstract} and \texttt{title\_abstract} modes and are truncated; the rate rises to 15.1\% for Statins, 10.4\% for Opioids, and 11.8\% for ADHD in the \texttt{title\_abstract\_mesh} mode, where the MeSH terms appended at the end of the input are most affected by truncation. The bag-of-words pipeline is unbounded and sees the full input in every case. This representation asymmetry is discussed in Sections~\ref{sec:discussion-design} and~\ref{sec:limitations}.

Class weights followed sklearn's balanced formulation, which is necessary at the inclusion rates the Cohen topics present. Without balanced weighting, BiomedBERT produced near-random discrimination (AUC $\approx$ 0.43) on the most imbalanced topic in pilot validation. Additional implementation details (mixed-precision training, seed argument mechanics, per-fold output format) are given in Appendix~\ref{sec:reproducibility}.

The choice of BiomedBERT rather than BioLinkBERT, ClinicalBERT, or BiomedBERT-large is deliberate. Our aim is not to identify the strongest screening architecture but to compare annotation-mechanism sensitivity across a bag-of-words baseline and a domain-pretrained contextual transformer at tractable fine-tuning cost. BiomedBERT-base places above BioBERT on BLURB \cite{gu2021biomedbert} at comparable parameter count, and the abstract-pretrained variant matches our input domain. Whether the finding generalises to other transformer architectures is an open question flagged in Section~\ref{sec:limitations}.

\subsection{Evaluation}

We report WSS@95\%, the work saved over sampling at 95\% recall \cite{cohen2006drug}, as the primary metric. WSS@95\% measures the fraction of the corpus a reviewer would not have to read after triaging by classifier-ranked probability, conditional on achieving 95\% recall of included articles. This metric is operational rather than statistical: it reflects the workload reduction a systematic review team would achieve if they used the classifier output to prioritise screening. Accuracy and ROC AUC are recorded per fold in the archived outputs but are not reported here, because they treat the two error directions symmetrically while screening practice does not.

Five-fold stratified cross-validation was applied to each topic in the main comparison. Stratification preserves the inclusion rate across folds, which matters at the 2.4\% inclusion rate of Opioids where a non-stratified split would risk folds with no positive examples.

\subsection{Statistical analysis and reproducibility characterisation}\label{sec:stats}

The per-fold WSS@95\% values from each topic-mode combination form the unit of analysis. For each topic and each classifier, the per-fold paired difference between expert mode and auto mode (\texttt{title\_abstract\_mesh} minus \texttt{auto\_mesh}) gives the observation set. Bag-of-words observations under the main protocol comprise seven reruns of five folds each, producing 35 paired differences per topic and 105 pooled across the three topics. BiomedBERT observations comprise five seeds of five folds each, producing 25 paired differences per topic and 75 pooled across the three topics.

Bootstrap percentile confidence intervals were computed with 10,000 resamples of the per-fold differences with replacement. The 2.5th and 97.5th percentiles of the resampled means give the 95\% CI. Paired permutation tests on the per-fold differences enumerate sign-flip patterns exactly for small samples and approximate by sampling for larger ones \cite{phipson2010permutation}. Naive paired $t$-tests are not reported: on $k$-fold CV data with overlapping training folds they under-estimate the variance of the mean difference \cite{nadeau2003inference}.

\textbf{Dependence structure and statistical unit.} The per-fold paired differences are the unit of analysis. Because folds share training data --- five folds share four fifths of it within a rerun, and all folds share the full corpus across reruns --- these differences are positively correlated. Resampling them as if they were independent understates the variance of their mean, so the intervals we report are narrower than the dependence warrants rather than wider, and we report them as descriptive summaries of uncertainty rather than as decision procedures. We compare intervals at several points below; in every case the comparison describes where two intervals lie, and we do not test the difference between two estimates, which would in any case require accounting for the fact that they are computed from the same corpus. 
Multi-run and multi-seed characterisation additionally treats each topic-classifier combination as a distribution. The two units answer different questions: the per-fold differences describe variation within the cross-validation procedure, while the run-level and seed-level means describe how far a repetition of the whole experiment moves. Section~\ref{sec:bow-results} reports the run-level distribution alongside the fold-level values so that both are visible, and run-level and seed-level means bound what a single-run point estimate can conceal.

\textbf{Non-determinism characterisation, motivating the multi-run and multi-seed protocols.} A reproducibility audit conducted earlier in this work surfaced two non-determinism findings that motivated the protocols this paper uses. The bag-of-words pipeline produces different per-fold WSS values across identical-command reruns, with a maximum per-fold spread of 0.144 WSS@95\% across the seven Statins reruns, and two BiomedBERT analyses on Statins with different random-state initialisations gave per-fold values that drifted by up to 0.26 WSS@95\%. The candidate sources (Keras layer initialisation under the current TensorFlow build; numpy RNG state at fine-tuning start) are documented in Appendix~\ref{sec:reproducibility}. What matters here is the methodological consequence: point-estimate reporting can conceal both run-to-run and seed-to-seed variation of magnitudes that exceed the effect sizes under discussion. The multi-run bag-of-words and multi-seed BiomedBERT protocols reported in Sections~\ref{sec:bow-results} and~\ref{sec:bert-results} treat each topic-classifier combination as a distribution. The same recognition prompted the design-sensitivity analysis: if concealed variation exceeded the effect size, so might variation across evaluation-design choices.

\subsection{Robustness analyses on the Statins finding}\label{sec:robustness}

The main comparison adopts the canonical Cohen evaluation design (5-fold stratified cross-validation at each topic's native corpus size). The Statins canonical result --- an expert-vs-auto MeSH gap that does not arise on Opioids or ADHD --- raises two questions the main design cannot answer. First, is the Statins effect a function of the larger corpus that Statins has relative to Opioids ($1.6\times$) and ADHD ($3.4\times$), rather than a topic-intrinsic property? Second, does the 5-fold fold count interact with the effect? The 5-fold design trains each model on 80\% of the topic's articles; the alternative 10-fold design trains on 90\%. We ran two robustness analyses to address each question separately.

The \textbf{matched-corpus-size analysis} stratified-subsampled Statins to $n=803$ articles, matching the ADHD corpus size. Seven subsample seeds were drawn, each preserving the Statins inclusion rate (5.5\%); each subsample was then run through the standard 5-fold pipeline with all four text modes, producing $7 \times 5 = 35$ per-fold paired differences on the subsampled Statins.

The \textbf{10-fold sensitivity analysis} retained Statins at full $n = 2{,}744$ but switched the cross-validation design from 5-fold to 10-fold, with seven reruns to match the existing multi-run protocol structure. This produces $7 \times 10 = 70$ per-fold paired differences. We did not run 10-fold on Opioids or ADHD because per-fold included-article counts at those topics under 10-fold ($\approx 4$ for Opioids, $\approx 8$ for ADHD) are too low for WSS@95\% to behave stably.

We also computed an \textbf{empirical power analysis} on the per-fold variances from a single 5-fold run of each topic under the canonical design. Those three runs were conducted in one earlier session, before the multi-run protocol was adopted. They are reported because a single canonical run is the design the screening literature has generally used, and the question is what such a design can detect: at the per-topic standard deviations a single canonical run produces, what is the minimum detectable effect (MDE) at 80\% power for a 5-fold design? Section~\ref{sec:design-sensitivity} reports the answer and what changes when the fold count of the multi-run protocol is substituted.

\section{Results}

\subsection{Bag-of-words results: topic-stratified multi-run characterisation}\label{sec:bow-results}

Tables~\ref{tab:bow_statins_multirun}, \ref{tab:bow_opiods_multirun}, and \ref{tab:bow_adhd_multirun} report per-fold expert-vs-auto MeSH WSS@95\% differences for the bag-of-words logistic regression classifier across all three topics, each characterised across seven reruns of the same pipeline configuration. The seven reruns per topic use the same code, input data, and fold-splitting seeds. Two of the seven Statins runs (runs 1 and 2) were executed with oneDNN kernels disabled, as part of the falsification test described in Appendix~\ref{sec:reproducibility}; the remaining five and all runs at Opioids and ADHD used the default build configuration. We report the seven together and note the difference here rather than treat the set as homogeneous: the run-level gaps from the two oneDNN-disabled runs ($+0.078$ and $+0.089$) fall inside the range of the other five ($[+0.077, +0.114]$). The run-to-run variation is consistent with residual non-determinism in Keras Dense layer initialisation that the pipeline's explicit seeding does not fully control, documented in Section~\ref{sec:stats}.

\begin{table*}[t]
\centering
\small
\begin{tabular}{rrrlr}
\toprule
Run & Expert WSS mean & Auto WSS mean & Per-fold gaps & Run gap \\
\midrule
1 & 0.204 & 0.126 & [+0.089, +0.027, +0.155, +0.000, +0.120] & +0.078 \\
2 & 0.198 & 0.109 & [+0.082, +0.004, +0.155, +0.073, +0.131] & +0.089 \\
3 & 0.218 & 0.142 & [+0.089, $-$0.053, +0.206, $-$0.022, +0.164] & +0.077 \\
4 & 0.200 & 0.098 & [+0.022, +0.024, +0.191, +0.148, +0.124] & +0.102 \\
5 & 0.213 & 0.115 & [+0.080, +0.027, +0.180, +0.062, +0.144] & +0.099 \\
6 & 0.216 & 0.105 & [+0.086, +0.055, +0.158, +0.111, +0.144] & +0.111 \\
7 & 0.227 & 0.113 & [+0.054, +0.127, +0.160, +0.100, +0.131] & +0.114 \\
\midrule
\textbf{Mean} & \textbf{0.211} & \textbf{0.115} & --- & \textbf{+0.096} \\
\bottomrule
\end{tabular}
\caption{BoW Statins multi-run characterisation, canonical 5-fold full-corpus design. Seven reruns of the same pipeline configuration; runs 1 and 2 were executed with oneDNN kernels disabled, as noted in the text.}
\label{tab:bow_statins_multirun}
\end{table*}

\begin{table*}[t]
\centering
\small
\begin{tabular}{rrrlr}
\toprule
Run & Expert WSS mean & Auto WSS mean & Per-fold gaps & Run gap \\
\midrule
1 & 0.277 & 0.279 & [$-$0.296, +0.025, $-$0.077, +0.274, +0.065] & $-$0.002 \\
2 & 0.267 & 0.267 & [$-$0.262, +0.023, +0.087, $-$0.048, +0.201] & +0.000 \\
3 & 0.310 & 0.303 & [$-$0.304, +0.098, $-$0.070, +0.156, +0.155] & +0.007 \\
4 & 0.295 & 0.288 & [$-$0.183, +0.031, $-$0.045, +0.139, +0.090] & +0.006 \\
5 & 0.295 & 0.257 & [$-$0.214, +0.022, +0.057, +0.200, +0.125] & +0.038 \\
6 & 0.318 & 0.302 & [$-$0.313, +0.073, +0.026, +0.082, +0.212] & +0.016 \\
7 & 0.306 & 0.325 & [$-$0.411, +0.040, $-$0.051, +0.184, +0.141] & $-$0.019 \\
\midrule
\textbf{Mean} & \textbf{0.295} & \textbf{0.289} & --- & \textbf{+0.007} \\
\bottomrule
\end{tabular}
\caption{BoW Opioids multi-run characterisation, canonical 5-fold full-corpus design.}
\label{tab:bow_opiods_multirun}
\end{table*}

\begin{table*}[t]
\centering
\small
\begin{tabular}{rrrlr}
\toprule
Run & Expert WSS mean & Auto WSS mean & Per-fold gaps & Run gap \\
\midrule
1 & 0.429 & 0.406 & [+0.168, $-$0.174, +0.180, +0.081, $-$0.138] & +0.023 \\
2 & 0.442 & 0.407 & [+0.254, $-$0.249, +0.075, +0.100, $-$0.006] & +0.035 \\
3 & 0.469 & 0.461 & [+0.354, $-$0.199, $-$0.006, +0.013, $-$0.119] & +0.009 \\
4 & 0.387 & 0.393 & [+0.180, $-$0.255, +0.081, +0.037, $-$0.075] & $-$0.006 \\
5 & 0.408 & 0.436 & [+0.143, $-$0.174, $-$0.013, +0.050, $-$0.144] & $-$0.028 \\
6 & 0.434 & 0.391 & [+0.267, $-$0.162, +0.099, +0.057, $-$0.044] & +0.043 \\
7 & 0.360 & 0.395 & [+0.217, $-$0.310, $-$0.056, +0.062, $-$0.087] & $-$0.035 \\
\midrule
\textbf{Mean} & \textbf{0.418} & \textbf{0.413} & --- & \textbf{+0.006} \\
\bottomrule
\end{tabular}
\caption{BoW ADHD multi-run characterisation, canonical 5-fold full-corpus design.}
\label{tab:bow_adhd_multirun}
\end{table*}

Under the canonical evaluation design, the Statins distribution is tight and positive: per-run mean $+0.096$, standard deviation 0.015 across seven reruns (standard error of the mean approximately 0.006), range $[+0.077, +0.114]$, all seven run-level gaps positive, 32 of 35 fold-level paired differences strictly positive. The two negative folds at Statins (Run 3 fold 2 at $-0.053$, Run 3 fold 4 at $-0.022$) are small in magnitude relative to the per-fold WSS standard deviation of approximately 0.07 across Statins folds.

Two earlier single runs of the same pipeline, conducted before the multi-run protocol was adopted, gave Statins gaps of $+0.121$ and $+0.125$. Both lie above the observed multi-run range upper bound of $+0.114$. They are consistent with single realisations from the upper part of the distribution the seven reruns describe, and the observation that a single run can land there is what prompted multi-run characterisation in the first place.

The Opioids and ADHD distributions are not positive and not tight. On Opioids the per-run mean is $+0.007$ with per-run range $[-0.019, +0.038]$, five of seven runs positive, 23 of 35 folds positive. On ADHD the per-run mean is $+0.006$ with per-run range $[-0.035, +0.043]$, four of seven runs positive, 18 of 35 folds positive. The per-fold gaps span both signs widely: Opioids fold gaps range from $-0.411$ to $+0.274$, ADHD fold gaps range from $-0.310$ to $+0.354$. Per-fold standard deviations within topic are 0.171 (Opioids) and 0.162 (ADHD), which is approximately 25 times the magnitudes of the run-level mean gaps. Under the canonical design, the gap is not detectable in either direction at these two topics.

\subsection{Each augmentation against its own baseline}\label{sec:decomposition}

Modes 1 and 2 provide a no-MeSH baseline for each of the two augmented modes: \texttt{title\_abstract} for the expert mode, which appends terms to title plus abstract, and \texttt{abstract} for the mechanical mode, which appends to the abstract alone. Measuring each augmentation against its own baseline removes the title from the comparison and separates what each assignment mechanism contributes. Table~\ref{tab:decomposition} reports the result on the same seven Statins reruns, using the same per-fold unit and bootstrap procedure as Section~\ref{sec:stats}.

\begin{table}[htbp]
\centering
\small
\begin{tabular}{lrr}
\toprule
Comparison & Mean & 95\% CI \\
\midrule
Expert increment (mode 3 $-$ mode 2)      & $+0.084$ & $[+0.067, +0.100]$ \\
Mechanical increment (mode 4 $-$ mode 1)  & $-0.006$ & $[-0.023, +0.012]$ \\
Difference between increments              & $+0.090$ & $[+0.062, +0.116]$ \\
Direct comparison (mode 3 $-$ mode 4)      & $+0.096$ & $[+0.075, +0.116]$ \\
\bottomrule
\end{tabular}
\caption{Each MeSH augmentation measured against its own no-MeSH baseline, Statins, seven reruns of 5-fold cross-validation, $n = 35$ per-fold differences per row. The final row is the direct expert-versus-auto comparison reported in Table~\ref{tab:bow_statins_multirun}, repeated for reference.}
\label{tab:decomposition}
\end{table}

The expert augmentation adds $+0.084$ WSS@95\% over its own baseline, with an interval that excludes zero and all seven run-level values positive. The mechanical augmentation adds $-0.006$, with an interval that spans zero: at the sample size available, appending mechanically-matched MeSH terms to the abstract has no detectable effect on WSS@95\% relative to the abstract alone. The title itself contributes $+0.006$ on average across the seven runs, positive in five of them.

Two things follow. The canonical Statins gap is not an artefact of the title asymmetry between modes 3 and 4: it stands at $+0.090$ when each augmentation is measured against its own baseline, against $+0.096$ under the direct comparison. And the gap arises because expert assignment adds signal rather than because mechanical assignment removes it. We describe the mechanical increment as not detectable rather than as negative, since its interval includes zero.

\subsection{What mechanical assignment recovers}\label{sec:assignment}

Section~\ref{sec:decomposition} shows that mechanical assignment contributes no detectable signal over the abstract alone. To characterise why, we compared the two term sets directly for each Statins article. Because the matcher tests whether a vocabulary term occurs as a substring of the abstract, an expert term it fails to recover is by construction a term whose surface form does not appear anywhere in that abstract.

Table~\ref{tab:assignment} reports the comparison, stratified by inclusion
label. Substring matching recovers roughly a sixth of the expert-assigned
terms, and about three quarters of what it does match was not part of the
expert annotation for that article. That the recovery rates for included and
excluded articles are close is the observation that bears on WSS@95\%: the
mechanical augmentation does not distinguish included from excluded articles,
so it adds little that a ranking classifier can use.

\begin{table}[htbp]
\centering
\small
\begin{tabular}{lrr}
\toprule
 & Included & Excluded \\
 & ($n=152$) & ($n=2{,}592$) \\
\midrule
Expert terms assigned per article        & 20.5  & 19.0  \\
Vocabulary terms found in abstract       & 14.1  & 13.1  \\
\quad of which expert-assigned           & 3.3   & 2.9   \\
\quad of which not assigned              & 10.8  & 10.1  \\
\midrule
Recovery rate                            & 16.2\% & 15.5\% \\
Unassigned share of matched terms        & 76.4\% & 77.6\% \\
\bottomrule
\end{tabular}
\caption{Expert versus mechanical MeSH assignment on Statins, per article,
stratified by inclusion label. Recovery rate is the share of expert-assigned
terms whose surface form appears in the abstract. Each row is a mean over
articles, rounded independently, so the two component rows need not sum exactly
to the row above them and the derived rates are computed before rounding.}
\label{tab:assignment}
\end{table}

The terms the matcher misses fall into recognisable classes. The most frequently missed are check tags recording study population rather than article content (\emph{humans}, missed in 148 of the 152 included articles, followed by \emph{male}, \emph{female}, \emph{middle aged} and \emph{aged}), and subheadings that an indexer attaches to a heading rather than writing into prose (\emph{therapeutic use}, \emph{drug therapy}, \emph{blood}, \emph{prevention \& control}). Neither class is expressible in abstract prose, so no surface method could recover them. A third class is more informative for the present question: pharmacological and structural class terms assigned to articles that name only a specific drug, such as \emph{anticholesteremic agents} (97 articles), \emph{hydroxymethylglutaryl-CoA reductase inhibitors} (58), and \emph{heptanoic acids} and \emph{pyrroles} (39 each). A fourth is lexicalisation mismatch, where the concept is present in the text but not in the MeSH surface form, as with \emph{double-blind method} (53).

The terms the matcher adds without expert warrant are dominated by general vocabulary (\emph{patients}, \emph{disease}, \emph{cholesterol}, \emph{risk}) together with artefacts of unbounded substring matching, in which \emph{back} matches inside \emph{background} and \emph{tics} inside \emph{statistics}.

We do not stratify recovery by term class quantitatively, so the 16\% figure should be read as a rate over all assigned terms rather than as a bound on what a surface method could in principle achieve; the first two classes are unrecoverable by any such method. The third and fourth classes are the ones that bear on Section~\ref{sec:lexical}.

\subsection{BiomedBERT results: multi-seed characterisation across three topics}\label{sec:bert-results}

BiomedBERT was characterised across five seeds (42, 7, 13, 21, 31) for each of the two modes central to RQ2 (\texttt{title\_abstract\_mesh} and \texttt{auto\_mesh}) on each of the three topics. This produces 25 per-fold WSS@95\% values per topic per mode (5 seeds $\times$ 5 folds), and 75 pooled across the three topics per mode. Tables~\ref{tab:bert_statins}, \ref{tab:bert_opiods}, and \ref{tab:bert_adhd} report the per-seed expert and auto WSS@95\% means and the per-seed expert-vs-auto gap for each topic.

\begin{table*}[t]
\centering
\small
\begin{tabular}{rrrr}
\toprule
Seed & Expert WSS mean & Auto WSS mean & Gap \\
\midrule
42 & 0.275 & 0.273 & +0.002 \\
7  & 0.256 & 0.253 & +0.004 \\
13 & 0.185 & 0.144 & +0.040 \\
21 & 0.342 & 0.282 & +0.060 \\
31 & 0.307 & 0.314 & $-$0.007 \\
\midrule
\textbf{Pooled (25 folds)} & \textbf{0.273} & \textbf{0.253} & \textbf{+0.020} \\
\bottomrule
\end{tabular}
\caption{BiomedBERT Statins multi-seed characterisation. Five seeds, five folds each.}
\label{tab:bert_statins}
\end{table*}

\begin{table*}[t]
\centering
\small
\begin{tabular}{rrrr}
\toprule
Seed & Expert WSS mean & Auto WSS mean & Gap \\
\midrule
42 & 0.367 & 0.441 & $-$0.073 \\
7  & 0.362 & 0.421 & $-$0.059 \\
13 & 0.438 & 0.357 & \textbf{+0.082} \\
21 & 0.402 & 0.529 & $-$0.127 \\
31 & 0.276 & 0.340 & $-$0.063 \\
\midrule
\textbf{Pooled (25 folds)} & \textbf{0.369} & \textbf{0.417} & \textbf{$-$0.048} \\
\bottomrule
\end{tabular}
\caption{BiomedBERT Opioids multi-seed characterisation. Five seeds, five folds each. Seed 13 produces a gap of $+0.082$ against the other four seeds' range of $[-0.127, -0.059]$. The outlier is retained without exclusion.}
\label{tab:bert_opiods}
\end{table*}

\begin{table*}[t]
\centering
\small
\begin{tabular}{rrrr}
\toprule
Seed & Expert WSS mean & Auto WSS mean & Gap \\
\midrule
42 & 0.601 & 0.656 & $-$0.055 \\
7  & 0.578 & 0.547 & +0.031 \\
13 & 0.606 & 0.544 & +0.062 \\
21 & 0.618 & 0.610 & +0.008 \\
31 & 0.639 & 0.667 & $-$0.029 \\
\midrule
\textbf{Pooled (25 folds)} & \textbf{0.608} & \textbf{0.605} & \textbf{+0.003} \\
\bottomrule
\end{tabular}
\caption{BiomedBERT ADHD multi-seed characterisation. Five seeds, five folds each.}
\label{tab:bert_adhd}
\end{table*}

Under the canonical 5-fold full-corpus evaluation, the BiomedBERT Statins per-seed gap distribution centres above zero (pooled $+0.020$, per-seed range $[-0.007, +0.060]$), but the central tendency is substantially below the bag-of-words Statins multi-run mean of $+0.096$. The Opioids per-seed gap distribution has a negative central tendency ($-0.048$ pooled, four of five seeds negative) with one seed-level outlier (seed 13 at $+0.082$) that we report rather than exclude. The ADHD per-seed gap distribution is centred near zero (pooled $+0.003$, three seeds positive of which one is within $0.01$ of zero, two negative).

The seed 13 Opioids result deserves explicit comment. Across five seeds with identical hyperparameters, fold splits, and training data, the per-seed gap ranges from $-0.127$ (seed 21) to $+0.082$ (seed 13), a swing of $0.21$ WSS@95\% attributable to random initialisation alone. This seed-level swing is of the same order as the single-run minimum detectable effect at Opioids that Section~\ref{sec:design-sensitivity} reports ($0.254$), meaning that seed-level instability at Opioids is comparable in magnitude to the smallest effect a single canonical run could detect there. The four-seed mean excluding seed 13 would be $-0.081$; the five-seed pooled mean retaining seed 13 is $-0.048$. We retain seed 13 because excluding it would amount to picking the result from a five-seed distribution post-hoc, and the wider distribution itself is part of the methodological observation: at this corpus size and inclusion rate (Opioids has 43 included articles in 1,772), per-seed fluctuation is substantial enough that single-seed characterisation can mislead about both magnitude and direction.

\subsection{Statistical tests}\label{sec:stat-tests}

Table~\ref{tab:stats} reports the bootstrap CIs and paired permutation $p$-values for the expert-minus-auto difference, per topic and pooled, separately for each classifier family, under the canonical 5-fold full-corpus design.

\begin{table*}[t]
\centering
\small
\begin{tabular}{llrrcr}
\toprule
Classifier & Topic & $n$ folds & Mean diff & 95\% bootstrap CI & Perm $p$ \\
\midrule
BoW (multi-run)  & Statins         & 35  & $+0.0957$ & $[+0.075, +0.116]$ & $< 0.001$ \\
                 & Opioids         & 35  & $+0.0066$ & $[-0.050, +0.061]$ & $0.82$ \\
                 & ADHD            & 35  & $+0.0059$ & $[-0.046, +0.060]$ & $0.83$ \\
                 & \textbf{Pooled} & \textbf{105} & \textbf{$+0.0361$} & \textbf{$[+0.009, +0.063]$} & \textbf{$0.013$} \\
\midrule
BERT (multi-seed) & Statins        & 25  & $+0.020$  & $[-0.021, +0.062]$ & $0.36$ \\
                 & Opioids         & 25  & $-0.048$  & $[-0.098, +0.004]$ & $0.08$ \\
                 & ADHD            & 25  & $+0.003$  & $[-0.040, +0.042]$ & $0.88$ \\
                 & \textbf{Pooled} & \textbf{75} & \textbf{$-0.008$} & \textbf{$[-0.036, +0.018]$} & \textbf{$0.55$} \\
\bottomrule
\end{tabular}
\caption{Statistical analysis of the expert$-$auto WSS@95\% difference by classifier and topic under the canonical 5-fold full-corpus design. Bootstrap CIs from 10,000-resample percentile bootstrap on the per-fold differences. Permutation $p$-values from sign-flip resampling at 100,000 permutations on the same unit. Intervals are descriptive, per Section~\ref{sec:stats}.}
\label{tab:stats}
\end{table*}

The bag-of-words Statins interval excludes zero and its permutation $p$ is below 0.001. At Opioids and ADHD it does not: both intervals include zero and both permutation $p$-values exceed 0.8. Pooling the three topics retains a small positive central tendency dominated by the Statins contribution and excludes zero on the lower bound of the interval by a small margin; the pooled permutation $p$-value of $0.013$ reflects the Statins component rather than a consistent cross-topic effect.

The BiomedBERT per-topic intervals all include zero. The Statins interval includes zero on the lower bound, the Opioids interval on the upper bound, and the ADHD interval is centred on zero. The three-topic pooled BiomedBERT interval of $[-0.036, +0.018]$ is centred essentially on zero and lies wholly below the bag-of-words Statins interval under the canonical design, which we report as a description of where the two intervals sit. At Opioids the seed-level instability described in Section~\ref{sec:bert-results} is the more informative reading of that topic.

\subsection{Evaluation design sensitivity}\label{sec:design-sensitivity}

\begin{figure*}[t]
\centering
\includegraphics[width=\textwidth]{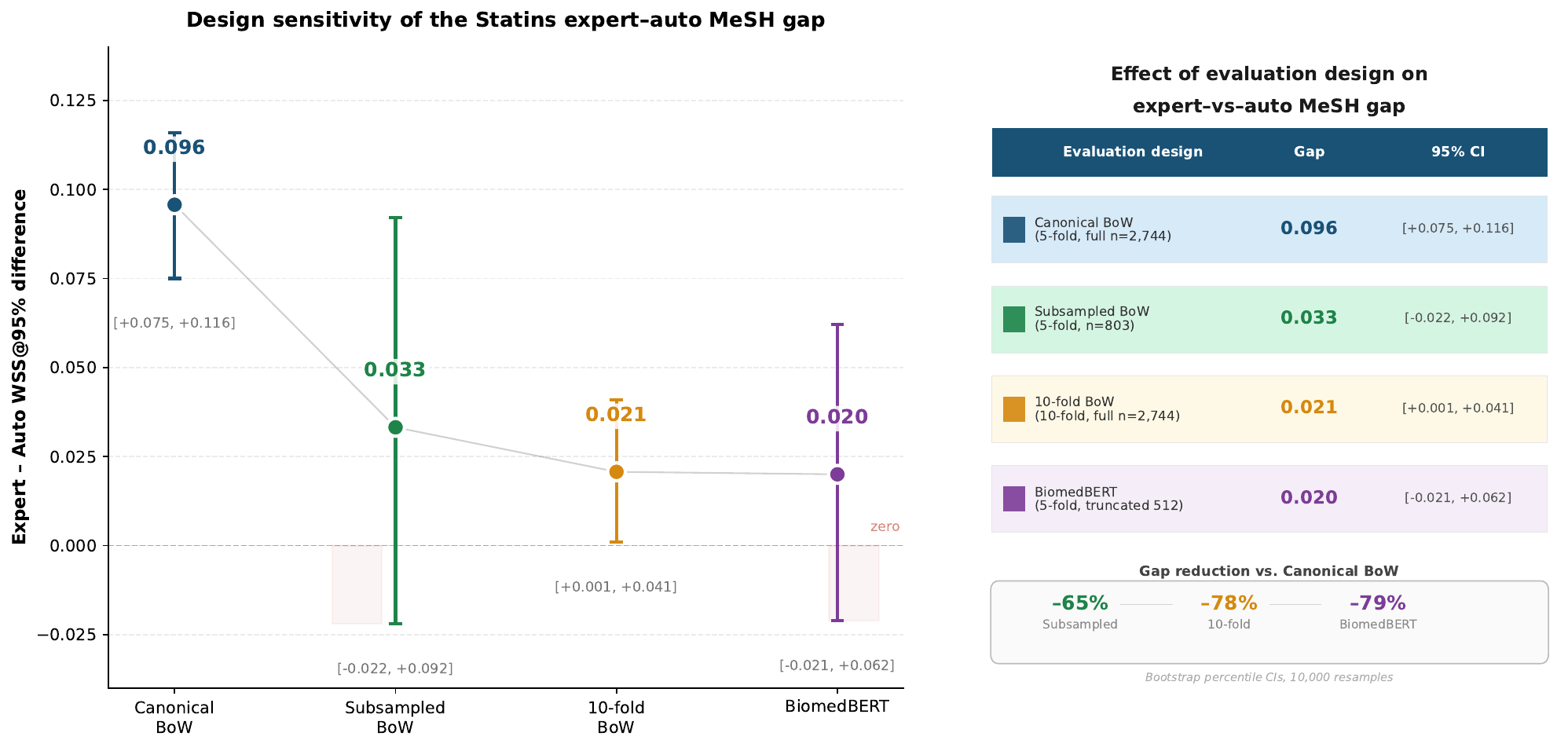}
\caption{Design sensitivity of the Statins expert--auto MeSH gap.
\textit{Left:} Mean paired WSS@95\% differences with 95\% bootstrap
percentile CIs (10,000 resamples). The canonical BoW result ($+0.096$)
attenuates under matched-corpus subsampling ($+0.033$, CI includes zero)
and 10-fold cross-validation ($+0.021$, CI excludes zero by a small
margin). The fourth column, BiomedBERT ($+0.020$, CI includes zero on the
lower bound), is not a change of evaluation design: it is the canonical
5-fold design with the classifier substituted, and its point estimate
differs from the BoW 10-fold result by $0.001$. Red shading indicates
CIs spanning zero. \textit{Right:} Summary table with gap magnitudes and,
for the three bag-of-words conditions, percentage change relative to the
canonical reference.}
\label{fig:design_sensitivity}
\end{figure*}

Two robustness analyses interrogate the Statins canonical finding by varying the evaluation design while holding the classifier, the data, and the assignment mechanism constant. Figure~\ref{fig:design_sensitivity} and Table~\ref{tab:design_sensitivity} report the results. The matched-corpus-size analysis subsamples Statins to $n = 803$ articles, matching ADHD's corpus size, and runs the standard 5-fold protocol across seven subsample seeds. The 10-fold sensitivity analysis retains Statins at full $n = 2{,}744$ but uses 10-fold instead of 5-fold cross-validation across seven reruns.

\begin{table*}[t]
\centering
\small
\begin{tabular}{lrrrr}
\toprule
Design & $n$ folds & Mean gap & 95\% bootstrap CI & SD \\
\midrule
Statins, 5-fold, full $n = 2{,}744$ (canonical, reference)   & 35  & $+0.096$ & $[+0.075, +0.116]$  & 0.064 \\
\textbf{Statins, 5-fold, subsampled $n = 803$}                & \textbf{35}  & \textbf{$+0.033$} & \textbf{$[-0.022, +0.092]$} & \textbf{0.177} \\
\textbf{Statins, 10-fold, full $n = 2{,}744$}                  & \textbf{70}  & \textbf{$+0.021$} & \textbf{$[+0.001, +0.041]$} & \textbf{0.086} \\
Opioids, 5-fold, multi-run reference $n = 1{,}772$              & 35  & $+0.007$ & $[-0.050, +0.061]$  & 0.171 \\
ADHD, 5-fold, multi-run reference $n = 803$                     & 35  & $+0.006$ & $[-0.046, +0.060]$  & 0.162 \\
\bottomrule
\end{tabular}
\caption{Evaluation design sensitivity on the Statins expert-vs-auto MeSH gap. The canonical Statins finding under the standard Cohen design (top row) attenuates under both subsampling to ADHD-matched size (row 2) and under 10-fold evaluation at full size (row 3). The subsampled interval includes zero. The 10-fold interval is narrowly positive and the magnitude falls by roughly four fifths relative to the canonical reference. Opioids and ADHD reference rows are the multi-run means from Tables~\ref{tab:bow_opiods_multirun}--\ref{tab:bow_adhd_multirun}, matching the same $n = 35$ per-fold structure as the Statins subsampled and canonical rows for commensurability. SD is the standard deviation of the per-fold paired differences at the stated fold count.}
\label{tab:design_sensitivity}
\end{table*}

Three findings emerge. First, the Statins gap falls by roughly two thirds at matched corpus size ($+0.033$ against the canonical $+0.096$) and the 95\% bootstrap interval includes zero. The interval for subsampled Statins ($[-0.022, +0.092]$) overlaps that for full-size ADHD ($[-0.046, +0.060]$) across almost its whole range, which we report as a description of the two intervals rather than as a test that the two conditions are indistinguishable. Second, at full corpus size but with 10-fold cross-validation, the Statins gap falls by roughly four fifths to $+0.021$, with an interval that excludes zero only by a small margin ($[+0.001, +0.041]$). Third, the 10-fold Statins point estimate and the canonical BiomedBERT Statins point estimate differ by $0.001$ WSS@95\% ($+0.021$ against $+0.020$), and their intervals overlap across most of their range; we do not test the difference between them. Both are roughly a fifth of the canonical BoW 5-fold reference.

The bag-of-words classifier therefore produces a substantially different expert-vs-auto gap depending on two evaluation choices that prior work treats as methodological defaults: how the corpus is sized (native versus matched) and how the cross-validation is partitioned (5-fold versus 10-fold). The canonical Cohen design happens to be the one under which the gap is most cleanly detected. We do not claim that the canonical design is the wrong default; we report that the magnitude of the Statins finding is partly a function of this default rather than an intrinsic property of the topic-classifier-mechanism combination.

We also computed an empirical power analysis on the per-fold variances from a single canonical 5-fold run at each topic. Table~\ref{tab:power} reports the minimum detectable effect (MDE) at 80\% power for a design of that shape.

\begin{table*}[htbp]
\centering
\small
\begin{tabular}{lrrrr}
\toprule
Topic & $n$ folds & Observed mean & Per-fold SD & MDE (80\% power) \\
\midrule
Statins & 5  & $+0.125$ & 0.068 & 0.114 \\
Opioids & 5  & $-0.010$ & 0.151 & 0.254 \\
ADHD    & 5  & $-0.030$ & 0.228 & 0.384 \\
\bottomrule
\end{tabular}
\caption{Empirical power analysis on the per-fold variances of a single canonical 5-fold run at each topic. Those three runs were conducted in one session predating the multi-run protocol, which is why the fold count differs from Table~\ref{tab:design_sensitivity}; a single canonical run is the design whose detection capacity this table characterises. At Statins and Opioids the single-run per-fold SDs (0.068, 0.151) are close to the corresponding multi-run values (0.064, 0.171); at ADHD the single-run SD (0.228) exceeds the multi-run value (0.162). MDE computed at $\alpha = 0.05$ two-sided, 80\% power, from the exact noncentral-$t$distribution. The normal approximation gives 0.085, 0.189 and 0.286 respectively, about a quarter smaller, and is not used.}
\label{tab:power}
\end{table*}

At the Statins per-fold variance ($0.068$), a 5-fold design gives a minimum detectable effect of $0.114$ at 80\% power, below the observed single-run Statins effect of $+0.125$ though not by a wide margin. At Opioids' larger per-fold variance ($0.151$) the same design gives an MDE of $0.254$, twice the Statins effect size, and at ADHD's variance ($0.228$) an MDE of $0.384$, more than three times it. A Statins-sized effect at Opioids or ADHD would not have been detected by a single canonical run at those variances.

This bound is a property of the design analysed, not of the fold count reported elsewhere in this paper. The multi-run protocol pools 35 per-fold differences per topic, and computing the same quantity at that fold count and those standard deviations gives MDEs of $0.031$ (Statins), $0.084$ (Opioids) and $0.079$ (ADHD), all below the $+0.096$ multi-run Statins effect. That calculation treats the 35 differences as independent observations, which Section~\ref{sec:stats} states they are not: folds within a rerun share four fifths of their training data and all reruns share the corpus. The effective number of independent observations therefore lies somewhere between the two, and the present design does not determine it. What the cross-topic null at Opioids and ADHD bounds is consequently detection capacity under an evaluation whose effective sample size is itself unresolved, and the data do not allow us to distinguish between ``no effect at those topics'' and ``effect present but not detected here.'' Resolving this would require either an analysis that models the fold dependence explicitly or replication on independent corpora.

\subsection{Pattern across topics, classifiers, and evaluation designs}

\begin{figure}[t]
\centering
\includegraphics[width=0.90\linewidth]{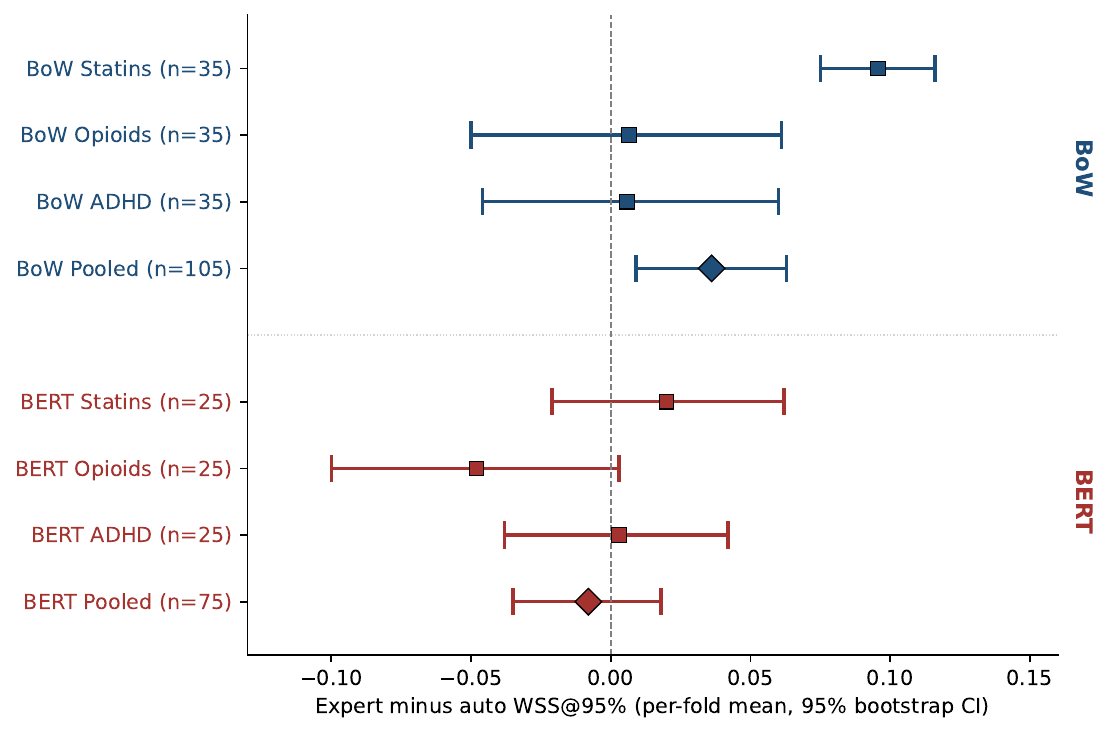}
\caption{Expert-minus-auto MeSH WSS@95\% differences across classifiers and topics, canonical design. Each row shows the per-fold mean difference (point) with a 95\% bootstrap percentile confidence interval (horizontal bar). BoW rows summarise seven-rerun multi-run characterisation (35 per-fold differences per topic, 105 pooled). BERT rows summarise five-seed multi-seed characterisation (25 per-fold differences per topic, 75 pooled). The BoW Statins interval $[+0.075, +0.116]$ and the BERT Statins interval $[-0.021, +0.062]$ do not overlap; we report this as a description of where the two intervals lie rather than as a test of the difference between the two estimates. Both classifiers show confidence intervals spanning zero at Opioids and ADHD. Figure~\ref{fig:design_sensitivity} shows the design-sensitivity analysis of the BoW Statins canonical finding.}
\label{fig:gap_forest}
\end{figure}

Figure~\ref{fig:gap_forest} summarises the main-comparison expert-minus-auto WSS@95\% gaps across the two classifier families and three topics. Under the canonical evaluation, the cross-classifier comparison admits a topic-stratified reading: where the bag-of-words classifier shows a positive expert-vs-auto MeSH gap with a tight distribution (Statins, $+0.096$, all seven reruns positive), BiomedBERT attenuates that gap to $+0.020$. Where the bag-of-words classifier shows no detectable gap (Opioids and ADHD), BiomedBERT shows the same null pattern.

The robustness analyses in Section~\ref{sec:design-sensitivity} require this reading to be qualified in two ways. First, the bag-of-words Statins gap is itself sensitive to two evaluation parameters; under both subsampling and 10-fold the gap attenuates to magnitudes close to the BiomedBERT canonical result. The transformer's apparent attenuation of the Statins gap, measured against the canonical BoW reference, becomes less distinctive when measured against the BoW result under designs that increase per-fold training volume. Second, the cross-topic null at Opioids and ADHD does not establish absence of a mechanism effect at those topics; at the single-run design the empirical MDE at their variances exceeds the Statins effect size, and at the pooled fold count the bound depends on an effective sample size the design does not determine.

The cleanest summary the data support is: the expert-vs-auto MeSH gap is present on Statins under the canonical Cohen evaluation design; under designs that increase per-fold training volume the gap shrinks to magnitudes close to those observed for BiomedBERT under the canonical design; and the cross-topic pattern at Opioids and ADHD is consistent with both ``no effect at those topics'' and ``effect present but not detected under the designs analysed.''

\section{Discussion}

\subsection{What this paper shows, design-conditional reading}\label{sec:discussion-design}

Under the canonical Cohen evaluation, expert MeSH outperforms mechanical MeSH by $+0.096$ WSS@95\% across seven reruns of the same pipeline configuration, with all seven run-level means positive and 32 of 35 fold-level paired differences positive. The difference persists across repetition of the pipeline at an evaluation design that is widely used in the field. The robustness analyses do not contradict that; they bound the conditions under which the difference has the magnitude reported.

They show that its magnitude depends on at least two evaluation design parameters the literature treats as defaults. Subsampling Statins to ADHD's corpus size reduces the gap by roughly two thirds and the interval includes zero. Switching from 5-fold to 10-fold at full corpus size reduces it by roughly four fifths, to $+0.021$. The 10-fold BoW Statins point estimate differs from the BiomedBERT 5-fold Statins point estimate by $0.001$. These observations are mutually consistent with two not-necessarily-exclusive readings.

The first reading is that the expert-vs-auto MeSH gap is a real and structural property of the topic, but that the classifier's response to the difference depends on how much training data the classifier has effectively absorbed. At smaller per-fold training volumes, the classifier relies on the MeSH context to a degree that makes the assignment-mechanism difference observable. At larger per-fold training volumes (either through 10-fold partitioning or through the representational capacity of a transformer), the classifier may extract more signal from the abstract alone, making the MeSH context less critical and shrinking the observed gap. This reading is a hypothesis the results are consistent with rather than a mechanism they isolate; the experiments do not measure what the classifier extracts. Under it, the 10-fold BoW result and the 5-fold BERT result may be explained by the same mechanism.

The alternative is that the canonical 5-fold full-corpus result is the point estimate that happens to be most cleanly recovered from the Cohen benchmark for Statins, and that the underlying effect is closer to the 10-fold magnitude. Under this reading, the canonical-design estimate overstates the effect's typical magnitude under other evaluation designs, and the present 10-fold result is an alternative point estimate the field could adopt. The intervals of the canonical and 10-fold designs do not overlap ($[+0.075, +0.116]$ against $[+0.001, +0.041]$); we do not formally test the difference between evaluation designs, and the two estimates are computed from the same corpus and are therefore not independent.

We do not adjudicate between these readings. Both are consistent with the data, and either would constitute a methodologically useful warning to the screening literature that the Cohen Statins effect is more design-conditional than has been recognised. Future work that varies per-fold training volume systematically (e.g.\ via learning-curve analysis at varying training-set fractions) could resolve the question.

A third contributing factor to the BoW-BERT difference at Statins, distinct from training volume, is BERT's 512-token truncation. In the \texttt{title\_abstract\_mesh} mode where the canonical bag-of-words gap arises, 15.1\% of Statins inputs are truncated at 512 BiomedBERT subword tokens (Section~\ref{sec:classifiers}), and the truncated content is preferentially the MeSH terms appended at the end of the input. The bag-of-words classifier sees the full MeSH augmentation in every case. Expert-assigned MeSH terms are typically more numerous than the mechanically-assigned terms produced by our substring-matching procedure, which is why the truncation rate in the expert-MeSH mode (15.1\%) is elevated relative to the abstract-only base rate (4.6--7.9\%). When truncation fires, expert mode therefore has more appended MeSH content available to lose than auto mode does. This asymmetry could attenuate the observed BiomedBERT gap, but establishing the direction empirically would require comparing performance on truncated against non-truncated examples, which we have not done: truncation changes the composition of what remains and not only its quantity. The contribution also cannot be isolated from the training-volume contribution without a no-truncation comparison (e.g.\ Longformer, or token-budget-restructured input ordering that places MeSH terms before the abstract body), which we leave as future work and disclose as a limitation in Section~\ref{sec:limitations}.

\subsection{One theoretical interpretation, not a demonstrated mechanism}\label{sec:lexical}

An earlier draft of this paper offered a lexical-semantic interpretation of the canonical Statins finding, drawing on the semasiological versus onomasiological distinction \cite{baldinger1980,geeraerts2010}. Mechanical MeSH lookup operates semasiologically: it begins with linguistic forms in the text and finds related forms in the controlled vocabulary. Expert MeSH annotation operates onomasiologically: the indexer begins with the article's topical content and selects vocabulary that expresses it.

We retain this as one interpretive frame, not as a demonstrated mechanism. Section~\ref{sec:assignment} provides one piece of descriptive evidence consistent with it: a substantial share of the expert terms mechanical matching cannot recover are class-level terms assigned to articles that name only a specific instance, and a further share are concepts present in the text but not in the MeSH surface form. That characterises the assignment difference, not the classifier's response to it. The distinction's observability in classifier outputs is in any case conditional on the classifier-evaluation combination: where the classifier has limited effective access to the abstract, the onomasiologically-curated annotation may carry more of the signal; where access is more effective, the abstract itself may supply what the augmentation would otherwise have provided. Direct evidence for this reading would require probing analysis of the transformer representations. The present experiments cannot distinguish the lexical-semantic interpretation from a purely training-volume interpretation, so we flag this as an open interpretive question rather than a claim the paper supports.

\subsection{Practical implication for screening pipelines}

The transformer's $+0.020$ Statins gap is in the same range as the BoW 10-fold Statins gap, and both are well below the canonical BoW 5-fold reference. Under the conditions tested, a screening pipeline using BiomedBERT with mechanical MeSH would see WSS@95\% performance within approximately $0.02$ of the expert-MeSH version on the Statins data; this is a small operational cost relative to the latency and coverage benefits of mechanical assignment. The same pipeline using bag-of-words logistic regression at 10-fold evaluation would show approximately the same cost. The classifier choice and the evaluation design therefore interact, and the canonical-design estimate of approximately $0.10$ WSS@95\% reported here is closer to a maximum than to a typical operational value.

This implication is design-conditional. Under the canonical Cohen evaluation, the gap a pipeline sees is the one reported in Section~\ref{sec:bow-results}. When the operational evaluation increases per-fold training volume, through 10-fold or through architectures that make more effective use of the abstract, the gap the pipeline sees is smaller. At topics where the expert-vs-auto gap is not detectable under any evaluation we tested, as at Opioids and ADHD in our data, the substitution question is moot. Expert annotation supports retrieval, browsing, and downstream uses for which mechanical assignment may not be adequate; the present implication concerns one specific downstream use (screening classifier training), rests on a single topic, and does not generalise beyond it without further evidence.

\subsection{What this finding does not establish}

The data do not establish that mechanical assignment outperforms expert assignment at the BiomedBERT scale on any of the three topics. They establish that expert assignment does not produce the magnitude of advantage predicted by the bag-of-words canonical-design Statins finding when the classifier is BiomedBERT or when the bag-of-words evaluation uses a design that increases per-fold training volume. Statements such as ``BERT reverses the gap'' or ``auto-MeSH outperforms expert MeSH'' would overstate what the analysis supports.

Nor does the analysis in Section~\ref{sec:decomposition} establish that mechanical assignment degrades performance. Its increment over the abstract-only baseline is $-0.006$ with an interval spanning zero, which supports the claim that mechanical assignment adds no detectable signal and does not support a claim that it removes signal.

The cross-topic null at Opioids and ADHD does not establish that no expert-vs-auto effect exists at those topics. At the single-run design the power analysis characterises, the empirical MDE at those topics' per-fold variances ($0.254$ for Opioids, $0.384$ for ADHD) exceeds the Statins effect size observed in the same analysis ($+0.125$), so a Statins-sized effect at either topic would have gone undetected. At the pooled fold count of the multi-run protocol the corresponding bound is smaller, and Section~\ref{sec:design-sensitivity} sets out why the effective sample size that decides between them is not determined by this design. Conclusions about whether the annotation-mechanism effect generalises across drug-class topics will require either larger samples, an analysis that models the fold dependence, or replication on independent corpora.

\subsection{What the finding leaves open for representation analysis and design isolation}

The present experiment shows that BiomedBERT under the canonical evaluation is approximately indifferent to the assignment mechanism on the topic where bag-of-words under the same evaluation shows the effect, and that bag-of-words under designs that increase per-fold training volume also shows the same approximate indifference. Whether this indifference reflects implicit recovery of the expert-curated structure inside the representation, an algorithmically alternative solution that achieves comparable performance without recovering that structure, or a saturation effect of training volume, cannot be determined from classification accuracy alone. Layer-wise probing of the hidden states, comparing recoverability of expert and auto MeSH structure across transformer layers and across designs where the assignment-mechanism effect varies in magnitude, is the natural next step. \newcite{plank2022} frames the broader concern about benchmark evaluation under label variation that this question connects to.

Two further design-isolation experiments would clarify the picture. A learning-curve analysis at varying training-set fractions of Statins (e.g.\ 25\%, 50\%, 75\%, 100\% with 5-fold) would isolate the per-fold training volume effect cleanly. A matched-evaluation-design comparison running BiomedBERT at both 5-fold and 10-fold on Statins would test whether the transformer effect is itself further attenuated by additional training data, or whether it has already saturated at 5-fold. Either is a natural extension of the present work.

\section{Limitations}\label{sec:limitations}

The experimental design has limitations that bound the interpretation of the result.

The transformer comparison uses a single architecture, BiomedBERT-base. Whether the finding generalises to BioBERT, BiomedBERT-large, BioLinkBERT, or to more recent biomedical large language models is an open question. The pattern's robustness across architectures is the natural next test.

The BiomedBERT 512-token sequence cap truncates a topic-dependent share of inputs: in the \texttt{title\_abstract\_mesh} mode, 15.1\% of Statins, 10.4\% of Opioids, and 11.8\% of ADHD inputs exceed the cap, and truncation preferentially removes the MeSH terms appended at the end of the input. The bag-of-words pipeline is unbounded and sees the full input in every case. This representation asymmetry could contribute to the smaller BiomedBERT Statins gap relative to the canonical bag-of-words reference; we cannot isolate this contribution from the training-volume contribution without a no-truncation BERT comparison (e.g.\ Longformer architecture or token-budget-restructured input ordering that places MeSH terms before the abstract body). We disclose this asymmetry in Section~\ref{sec:classifiers} rather than treat it as a confound the present analysis can correct for. The single-ontology constraint is similar: we tested MeSH only, and whether the same finding holds for other biomedical controlled vocabularies, such as UMLS, is not established by this experiment. The English-only scope is a third constraint that follows from the Cohen benchmark itself.

Three of fifteen available Cohen topics provide the cross-topic evidence. This is reasonable coverage but not exhaustive. Extending to the remaining twelve topics would clarify the topic-dependence finding.

The evaluation design sensitivity revealed by the robustness analyses bounds the generality of the canonical-design result. The two analyses are separate one-factor perturbations of the canonical design rather than components of a decomposition: matched-$n$ subsampling reduces the gap from $+0.096$ to $+0.033$ and 10-fold at full corpus size reduces it to $+0.021$, but each perturbation changes more than the parameter it is named for. Subsampling changes the absolute number of included articles and the composition of the corpus as well as its size; moving from 5-fold to 10-fold changes the test-set size and the number of evaluation observations as well as the training fraction. We have not run the joint design (10-fold at matched $n$), because the resulting per-fold included-article count at $n = 803$ under 10-fold ($\approx 4$) is too low for WSS@95\% to behave stably. What the two analyses jointly support is that the canonical-design magnitude is sensitive to both corpus size and fold configuration, not an attribution of the attenuation to either in isolation.

The BiomedBERT hyperparameters were chosen as standard fine-tuning defaults rather than optimised for the Cohen topics. Light tuning of the learning rate or number of epochs may produce modest gains in absolute WSS@95\% but is unlikely to change the qualitative comparison between expert and auto modes since both use identical training procedures. We did not run BiomedBERT under 10-fold evaluation; whether the transformer effect attenuates further at 10-fold or has already saturated at 5-fold is an open question.

The bag-of-words pipeline exhibits residual non-determinism between identical-command reruns, with a maximum per-fold spread of 0.144 WSS@95\% across the seven Statins reruns; the candidate mechanisms and the mitigation approach are documented in Appendix~\ref{sec:reproducibility}. Whether seven reruns are sufficient to characterise the run-to-run distribution is itself an assumption: at the observed run-to-run standard deviation of $0.015$ WSS@95\% on the Statins gap, seven reruns give a standard error of the mean of approximately $0.006$, small relative to the effect sizes under discussion.

The mechanical MeSH assignment procedure uses case-insensitive substring matching against a MeSH vocabulary built from the cache of PubMed records for the benchmark. Three properties of that construction bound the result. The vocabulary is a subset of the official MeSH vocabulary, restricted to headings an indexer assigned somewhere in the benchmark, so it is larger than a fold-scoped vocabulary and smaller than the full thesaurus; a deployment matcher working from the thesaurus would have more terms available, not fewer. It spans more of the benchmark than the three topics studied, so it admits matches the indexer did not assign for a given article, and Section~\ref{sec:assignment} reports that share. And matching is unbounded containment with a four-character floor, so short terms match inside longer words. We do not establish which way these properties move the observed gap, and we therefore do not present the reported gap as a bound on what a different mechanical procedure would produce. Alternative procedures (MetaMap, ScispaCy, cTAKES, MedCAT) perform contextual disambiguation that substring matching does not, and whether the finding extends to these is an additional open question. For the same reason, ``mechanical assignment'' throughout this paper denotes the specific lexical matcher described in Section~\ref{sec:classifiers} and not automatic MeSH indexing in general.

The seed 13 result on Opioids deserves a methodological caveat. Four seeds produce gaps in $[-0.127, -0.059]$ and one (seed 13) produces $+0.082$. Whether this represents a single tail draw from a wide distribution or evidence of a bimodal distribution at Opioids cannot be resolved with five seeds. A future seed pool of 15 or more seeds at Opioids would resolve this. Retaining the outlier in the reported pooled mean ($-0.048$) is the conservative choice.

\section{Conclusion}

We tested whether the expert-vs-auto MeSH gap observed in bag-of-words systematic review screening on the Cohen benchmark is stable across evaluation-design perturbations and across classifier families. Under the canonical Cohen evaluation the Statins gap is positive and tight across seven reruns, while Opioids and ADHD show null gaps. An empirical power analysis interprets those nulls against what a single canonical run can detect at their per-fold variances. Measured against each mode's own no-MeSH baseline, the expert augmentation contributes $+0.084$ WSS@95\% and the mechanical augmentation contributes nothing detectable. Two robustness analyses on Statins show that the canonical bag-of-words finding is conditional on the evaluation design: subsampling to matched corpus size reduces the gap by roughly two thirds with an interval that includes zero, and 10-fold cross-validation at full corpus size reduces it by roughly four fifths, to a point estimate $0.001$ from the canonical BiomedBERT result.

The expert-vs-auto MeSH gap is therefore present at the canonical Cohen evaluation but evaluation-design-conditional in a way we did not find characterised in the screening literature. Where the gap is detectable under canonical evaluation, transformer classifiers and bag-of-words classifiers under designs that increase per-fold training volume produce approximately the same attenuated gap; where it is not detectable under any tested design, the substitution question is moot. Whether the attenuation pattern generalises across transformer architectures, ontologies, and the remaining Cohen topics is an open question that Section~\ref{sec:limitations} enumerates in full. The broader contribution is methodological: benchmark conclusions about feature-source effects can change substantially under reasonable evaluation-design changes, even when the underlying task and data are held constant.

\bibliography{nejlt}

@article{cohen2006drug,
  author = {Cohen, Aaron M. and Hersh, William R. and Peterson, Kim and Yen, Po-Yin},
  title = {Reducing workload in systematic review preparation using automated citation classification},
  journal = {Journal of the American Medical Informatics Association},
  volume = {13},
  number = {2},
  pages = {206--219},
  year = {2006}
}

@inproceedings{cohen2008,
  author = {Cohen, Aaron M.},
  title = {Optimizing feature representation for automated systematic review work prioritization},
  booktitle = {{AMIA} Annual Symposium Proceedings},
  pages = {121--125},
  year = {2008}
}

@article{matwin2010,
  author = {Matwin, Stan and Kouznetsov, Alexandre and Inkpen, Diana and Frunza, Oana and O'Blenis, Peter},
  title = {A new algorithm for reducing the workload of experts in performing systematic reviews},
  journal = {Journal of the American Medical Informatics Association},
  volume = {17},
  number = {4},
  pages = {446--453},
  year = {2010}
}

@article{mao2017meshnow,
  author = {Mao, Yuqing and Lu, Zhiyong},
  title = {{MeSH Now}: automatic {MeSH} indexing at {PubMed} scale via learning to rank},
  journal = {Journal of Biomedical Semantics},
  volume = {8},
  number = {1},
  pages = {15},
  year = {2017}
}

@inproceedings{jin2018pico,
  author = {Jin, Qiao and Szolovits, Peter},
  title = {{PICO} element detection in medical text via long short-term memory neural networks},
  booktitle = {Proceedings of the {BioNLP} Workshop},
  pages = {67--75},
  year = {2018}
}

@inproceedings{scott1999,
  author = {Scott, Sam and Matwin, Stan},
  title = {Feature engineering for text classification},
  booktitle = {Proceedings of {ICML-99}},
  pages = {379--388},
  year = {1999}
}

@article{gu2021biomedbert,
  author = {Gu, Yu and Tinn, Robert and Cheng, Hao and Lucas, Michael and Usuyama, Naoto and Liu, Xiaodong and Naumann, Tristan and Gao, Jianfeng and Poon, Hoifung},
  title = {Domain-specific language model pretraining for biomedical natural language processing},
  journal = {{ACM} Transactions on Computing for Healthcare},
  volume = {3},
  number = {1},
  pages = {1--23},
  year = {2021}
}

@article{lee2020biobert,
  author = {Lee, Jinhyuk and Yoon, Wonjin and Kim, Sungdong and Kim, Donghyeon and Kim, Sunkyu and So, Chan Ho and Kang, Jaewoo},
  title = {{BioBERT}: a pre-trained biomedical language representation model for biomedical text mining},
  journal = {Bioinformatics},
  volume = {36},
  number = {4},
  pages = {1234--1240},
  year = {2020}
}

@inproceedings{yasunaga2022linkbert,
  author = {Yasunaga, Michihiro and Leskovec, Jure and Liang, Percy},
  title = {{LinkBERT}: pretraining language models with document links},
  booktitle = {Proceedings of the 60th Annual Meeting of the Association for Computational Linguistics ({ACL})},
  pages = {8003--8016},
  year = {2022}
}

@article{guo2024llm,
  author = {Guo, Eddie and Gupta, Mehul and Deng, Jiawen and Park, Ye-Jean and Paget, Michael and Naugler, Christopher},
  title = {Automated paper screening for clinical reviews using large language models: data analysis study},
  journal = {Journal of Medical Internet Research},
  volume = {26},
  pages = {e48996},
  year = {2024}
}

@article{alshami2023chatgpt,
  author = {Alshami, Ahmed and Elsayed, Moustafa and Ali, Eslam and Eltoukhy, Abdelrahman E. E. and Zayed, Tarek},
  title = {Harnessing the power of {ChatGPT} for automating systematic review process: methodology, case study, limitations, and future directions},
  journal = {Systems},
  volume = {11},
  number = {7},
  pages = {351},
  year = {2023}
}

@inproceedings{wang2024boolean,
  author = {Wang, Shuai and Scells, Harrisen and Koopman, Bevan and Zuccon, Guido},
  title = {Can {ChatGPT} write a good boolean query for systematic review literature search?},
  booktitle = {Proceedings of the 47th International {ACM} {SIGIR} Conference on Research and Development in Information Retrieval},
  pages = {1426--1436},
  year = {2024}
}

@inproceedings{plank2022,
  author = {Plank, Barbara},
  title = {The ``Problem'' of human label variation: on ground truth in data, modeling and evaluation},
  booktitle = {Proceedings of the 2022 Conference on Empirical Methods in Natural Language Processing ({EMNLP})},
  pages = {10671--10682},
  year = {2022}
}

@article{nadeau2003inference,
  author = {Nadeau, Claude and Bengio, Yoshua},
  title = {Inference for the generalization error},
  journal = {Machine Learning},
  volume = {52},
  number = {3},
  pages = {239--281},
  year = {2003}
}

@article{phipson2010permutation,
  author = {Phipson, Belinda and Smyth, Gordon K.},
  title = {Permutation {P}-values should never be zero: calculating exact {P}-values when permutations are randomly drawn},
  journal = {Statistical Applications in Genetics and Molecular Biology},
  volume = {9},
  number = {1},
  pages = {Article 39},
  year = {2010}
}

@book{baldinger1980,
  author = {Baldinger, Kurt},
  title = {Semantic Theory: Towards a Modern Semantics},
  publisher = {Blackwell},
  address = {Oxford},
  year = {1980}
}

@book{geeraerts2010,
  author = {Geeraerts, Dirk},
  title = {Theories of Lexical Semantics},
  publisher = {Oxford University Press},
  address = {Oxford},
  year = {2010}
}

@book{saussure1959,
  author = {de Saussure, Ferdinand},
  title = {Course in General Linguistics},
  publisher = {Philosophical Library},
  address = {New York},
  year = {1959},
  note = {Original French edition 1916}
}

@article{harris1954distributional,
  author = {Harris, Zellig S.},
  title = {Distributional structure},
  journal = {Word},
  volume = {10},
  number = {2--3},
  pages = {146--162},
  year = {1954}
}

@article{sahlgren2008,
  author = {Sahlgren, Magnus},
  title = {The distributional hypothesis},
  journal = {Italian Journal of Linguistics},
  volume = {20},
  number = {1},
  pages = {33--54},
  year = {2008}
}

@article{lenci2018,
  author = {Lenci, Alessandro},
  title = {Distributional models of word meaning},
  journal = {Annual Review of Linguistics},
  volume = {4},
  pages = {151--171},
  year = {2018}
}
\bibliographystyle{nejlt_bib}

\appendix

\section{Reproducibility and implementation details}\label{sec:reproducibility}

\subsection{Repository and per-fold outputs}

\newcommand{\repourl}{\url{https://github.com/SamInMotion/Medical-intervention-text-classification}, with an archived release at \texttt{10.5281/zenodo.21626925}}

The Cohen benchmark loader, BoW pipeline, BiomedBERT pipeline, evaluation metrics, the statistical analysis scripts, the robustness-analysis scripts, and the assignment analysis reported in Section~\ref{sec:assignment} are available at \repourl. The archived per-fold outputs permit exact reproduction of every table and figure in this paper; re-executing the pipelines produces small stochastic variation of the magnitude characterised in Section~\ref{sec:stats}, so a rerun reproduces the reported distributions rather than the individual per-fold values. Per-fold WSS@95\%, AUC, and accuracy values for all BiomedBERT runs are saved as \texttt{.json} sibling files and can be parsed deterministically. Multi-seed per-topic summaries, the seven-rerun bag-of-words summaries, and the per-fold values for the subsampling and 10-fold robustness analyses are all archived alongside them. A claim-to-source map naming the file behind every number in this paper is included in the repository.

\subsection{BiomedBERT implementation}

Mixed-precision training (fp16) was used on T4 GPUs to make the full four-mode comparison feasible within Google Colab's free-tier budget. The stratified fold splits and the \texttt{BertConfig.seed} are seeded explicitly through the \texttt{--seed} pipeline argument. Per-fold WSS values are written to \texttt{.json} sibling files alongside the standard \texttt{.txt} outputs for downstream parsing.

\subsection{Non-determinism sources}

The BoW pipeline uses nominally explicit seeds (\texttt{set\_seeds()} in \texttt{src/cohen\_pipeline.py}) which call \texttt{np.random.seed()} and \texttt{tf.random.set\_seed()}, but does not call \texttt{tf.keras.utils.set\_random\_seed()} or \texttt{tf.config.experimental.enable\_op\_determinism()}. Residual non-determinism in Keras Dense layer initialisation is the leading candidate for the consequence. The hypothesis that observed run-to-run drift is caused by oneDNN runtime optimisations was tested by running with \texttt{TF\_ENABLE\_ONEDNN\_OPTS=0} and falsified: two consecutive identical-command runs with oneDNN disabled still produced different per-fold values. That eliminates oneDNN as the sole source; it does not on its own isolate the mechanism, and the residual variation is consistent with Keras layer initialisation under the current TensorFlow build. Those two runs are retained in the multi-run characterisation and identified in Section~\ref{sec:bow-results}. Rather than pursue the mechanism further into the TensorFlow stack, we characterise the run-to-run distribution as reported there. The BiomedBERT pipeline uses explicit \texttt{--seed} arguments at each invocation and the multi-seed protocol disclosed in Section~\ref{sec:bert-results}; residual seed-to-seed variation of up to 0.26 WSS@95\% at Statins, and 0.28 at ADHD, observed in early runs motivated the five-seed characterisation across all three topics.

\newpage
\section{Hypothesis register state}

H2 in Section 1 was registered in the following pre-specified form prior to the BiomedBERT experiment:

\begin{quote}
\textbf{H-Pub2:} The bag-of-words finding (mechanical lookup fails, expert assignment succeeds) persists when the classifier is BiomedBERT. Specifically: BiomedBERT with auto-MeSH features will not recover the WSS@95\% gap relative to BiomedBERT with expert-MeSH features.

\textit{Falsifying evidence specified in advance:} BiomedBERT achieves comparable WSS@95\% with auto-MeSH features as with expert-MeSH features, with the gap closing to within standard deviations on at least Statins.
\end{quote}

The disposition is partial falsification, design-conditional. At Statins under the canonical evaluation, where the bag-of-words gap is positive and tight (multi-run mean $+0.096$, all seven reruns positive), BiomedBERT attenuates it to a multi-seed pooled $+0.020$. The reduction is substantial but does not close the gap to within seed-level standard deviations on the BERT side. The robustness analyses (Section~\ref{sec:design-sensitivity}) show that the bag-of-words canonical-design Statins gap itself attenuates substantially under matched-corpus-size and under 10-fold evaluation, with the 10-fold result ($+0.021$) differing from the canonical BERT result by $0.001$. The H-Pub2 framing of ``transformer absorbs a structural bag-of-words effect'' is therefore best read as ``transformer under canonical evaluation produces a result comparable to bag-of-words under designs that increase per-fold training volume''; both attenuate the canonical-design gap to a small positive magnitude that may or may not be different from zero depending on which interval one privileges. At Opioids and ADHD, where the bag-of-words gap is not detectable under any tested design and a single canonical run lacks the power to detect a Statins-sized effect at those topics' variances, H-Pub2 was not directly testable against the assumed-positive bag-of-words baseline. The framing supported by the evidence is therefore design-conditional attenuation at Statins, with the open question of whether the attenuation is a property of the transformer specifically or of the increased effective training volume that distinguishes BERT and BoW-10-fold from the BoW canonical baseline.

\end{document}